%% file: main.tex
\documentclass[10pt,twocolumn,letterpaper]{article}

\usepackage{cvpr}              
\input{preamble}
\definecolor{cvprblue}{rgb}{0.21,0.49,0.74}
\usepackage[pagebackref,breaklinks,colorlinks,allcolors=cvprblue]{hyperref}
\usepackage{xcolor}         
\def\paperID{} 
\def\confName{CVPR}
\def\confYear{2026}
\usepackage[table]{xcolor} 
\usepackage{bbding}
\usepackage[most]{tcolorbox}
\usepackage{pifont}
\newtcolorbox{promptbox}{
    colback=pink,           
    colframe=black,            
    arc=5pt,                   
    boxrule=1pt,               
    left=8pt, right=8pt,       
    top=6pt, bottom=6pt,       
    fonttitle=\bfseries,       
    title={Prompt for Tool Calling:}, 
    enhanced,                  
    drop shadow,               
}
\definecolor{indigo}{RGB}{75,0,130}
\newcommand*{\affaddr}[1]{#1} 
\newcommand*{\affmark}[1][*]{\textsuperscript{#1}}
\newcommand{\authormark}[2][]{%
  \begingroup
  \def\@thefnmark{#1}%
  \footnote{#2}%
  \endgroup
}
\title{CC-VQA: Conflict- and Correlation-Aware Method for \\ Mitigating Knowledge Conflict in Knowledge-Based Visual Question Answering}

\author{%
Yuyang Hong\affmark[1,2,3]\thanks{Equal contribution}~~\thanks{Work done during an internship at Alibaba Cloud}~~~ 
Jiaqi Gu\affmark[3]\footnotemark[1]~~~ 
Yujin Lou\affmark[3]~~~
Lubin Fan\affmark[3]\thanks{Corresponding author}~~~ 
Qi Yang\affmark[1,2]~~~  \vspace{3pt} \\
Ying Wang\affmark[2]\footnotemark[3]~~~
Kun Ding\affmark[2]~~~ 
Yue Wu\affmark[3]~~~ 
Shiming Xiang\affmark[1,2]~~~
Jieping Ye\affmark[3]~~~\vspace{3pt} \\
\affaddr{\affmark[1]School of Artificial Intelligence, UCAS~~~~~~~~~~}
\affaddr{\affmark[2]MAIS, Institute of Automation~~~~~~~~~~} 
 \\
\affaddr{\affmark[3]Alibaba Cloud Computing~~~~~~~~~~} \\
\small
}

\begin{document}
\maketitle
\input{sec/0_abstract}    
\input{sec/1_intro}
\input{sec/2_related_work}
\input{sec/3_observation}
\input{sec/4_method}

\input{sec/5_experiments}
\input{sec/6_conclusion}
{
    \small
    \bibliographystyle{ieeenat_fullname}
    \bibliography{main}
}
\input{sec/X_suppl}

\end{document}

%% file: sec/0_abstract.tex
\begin{abstract}
Knowledge-based visual question answering (KB-VQA) demonstrates significant potential for handling knowledge-intensive tasks. However, conflicts arise between static parametric knowledge in vision language models (VLMs) and dynamically retrieved information due to the static model knowledge from pre-training. The outputs either ignore retrieved contexts or exhibit inconsistent integration with parametric knowledge, posing substantial challenges for KB-VQA. Current knowledge conflict mitigation methods primarily adapted from language-based approaches, focusing on context-level conflicts through engineered prompting strategies or context-aware decoding mechanisms. However, these methods neglect the critical role of visual information in conflicts and suffer from redundant retrieved contexts, which impair accurate conflict identification and effective mitigation. To address these limitations, we propose \textbf{CC-VQA}: a novel training-free, conflict- and correlation-aware method for KB-VQA. Our method comprises two core components: (1) Vision-Centric Contextual Conflict Reasoning, which performs visual-semantic conflict analysis across internal and external knowledge contexts; and (2) Correlation-Guided Encoding and Decoding, featuring positional encoding compression for low-correlation statements and adaptive decoding using correlation-weighted conflict scoring. Extensive evaluations on E-VQA, InfoSeek, and OK-VQA benchmarks demonstrate that CC-VQA achieves state-of-the-art performance, yielding absolute accuracy improvements of 3.3\% to 6.4\% compared to existing methods. Code is available at \url{https://github.com/cqu-student/CC-VQA}.
\end{abstract}

%% file: sec/1_intro.tex
\section{Introduction}
\label{sec:intro}

Vision Language Models (VLMs)~\cite{achiam2023gpt,bai2025qwen2,zhu2025internvl3} demonstrate exceptional performance in Visual Question Answering (VQA)~\cite{antol2015vqa,goyal2017making} by leveraging extensive parametric knowledge learned during pre-training. However, limitations persist in Knowledge-Based Visual Question Answering (KB-VQA) due to static training data that cannot be dynamically updated and may contain incomplete information. While Retrieval-Augmented Generation (RAG) methods~\cite{yan2024echosight,cocchi2025augmenting,caffagni2024wiki,hongknowledge,qi2024rora} address this limitation by incorporating domain-specific knowledge without retraining, they introduce risks of knowledge conflicts between external knowledge and model-parametric knowledge. Figure 1 illustrates a knowledge conflict case in KB-VQA. Knowledge conflicts intensify as the knowledge base expands. The main reason is the unreliability of the knowledge predicted or retrieved by the model.

\begin{figure}[t]
    \centering
    \includegraphics[width=1\linewidth]{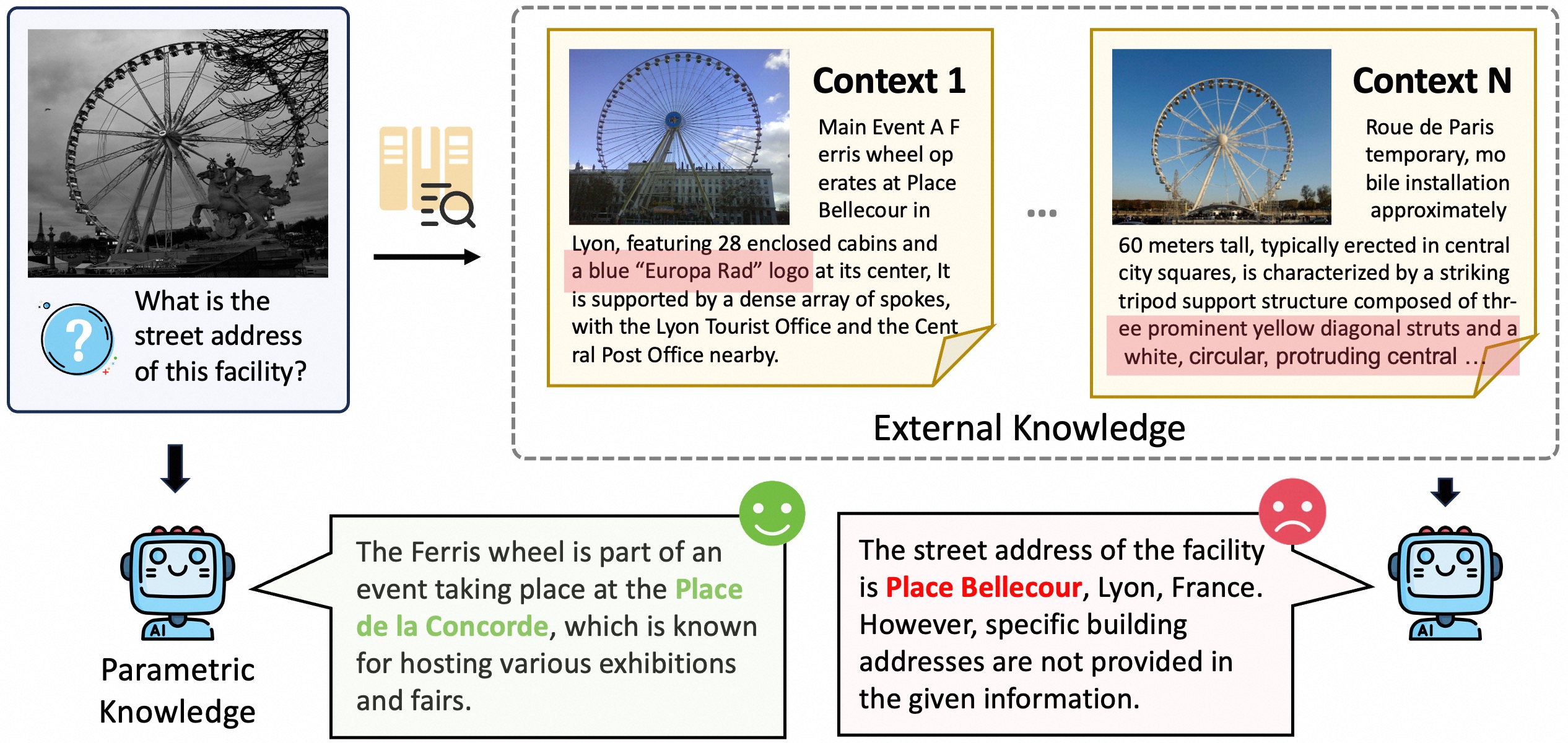}
    \caption{\textbf{Example of Knowledge Conflict in KB-VQA.} The introduction of external knowledge can enhance performance in knowledge-intensive visual question answering tasks; however, it concurrently introduces error risks stemming from conflicts between model parametric knowledge and external sources. Leveraging visual semantic features from both query image and contexts (highlighted in red) may mitigate knowledge conflicts and improve response accuracy in KB-VQA tasks.}
    \label{fig:knowledge_conflict}
    \vspace{-0.3cm}
\end{figure}

Knowledge conflicts were initially studied in knowledge-based language question answering (KB-QA)~\cite{xie2023adaptive,mingfaitheval,ying2024intuitive,yuan2024discerning}.
Recent studies reveal that retrieval-augmented generation (RAG) systems exhibit significant accuracy degradation under knowledge conflict conditions. The model will either default to its internal knowledge while ignoring the retrieved context, or get misled by contradictory retrieved contexts, which in turn increase the uncertainty of the final answer and ultimately degrading the overall performance. Current methods to mitigate conflicts in KB-QA can be categorized into two ways: (1) \textit{prompt-based methods}~\cite{zhang2025faithfulrag,zhou2023context,zhang2023merging} that optimize prompting strategies to align responses with contextual knowledge, and (2) \textit{decoding-based methods}~\cite{yuan2024discerning,wang2025adacad,khandelwal2025cocoa} that modify generation mechanisms through entropy constraints or contrastive decoding, enforcing closer integration of external context or internal knowledge during decoding.

Multimodal RAG systems face heightened complexity in knowledge conflicts, as visual inputs introduce additional information channels and potential inconsistencies. Beyond inheriting challenges from unimodal RAG systems, multimodal knowledge conflicts are further increased by cross-modal retrieval limitations, complex visual understanding requirements, and amplified model hallucination. However, few works have discussed the issue of knowledge conflict in the field of multimodal RAG systems.

Therefore, we propose a Conflict- and Correlation-Aware KB-VQA method, named \textbf{CC-VQA}, to mitigate knowledge conflicts in multi-modal RAG systems. Our method is motivated by two key principles: 
(1) vision-centric analysis reduces both visual ambiguity as well as reasoning uncertainty caused by conflicts;
(2) fine-grained analysis on contexts leads to sharper conflict detection and more accurate answers.
The CC-VQA framework comprises two components. First, the \emph{Contextual Conflict Reasoning} module externalizes model-parametric knowledge to form parametric contexts, then performs vision-centric conflict reasoning with retrieved knowledge. This stage explicitly reasons about inter-context conflicts associated with visual semantic features, generating enhanced contexts with conflict annotations. Second, the \emph{Correlation-Guided Encoding and Decoding} module conducts sentence-level correlation assessment across contexts. During encoding, positional encoding compression prioritizes low-correlation statements; during decoding, adaptive token sampling adjusts distributions based on correlation weights. This generation mechanism significantly improves answer accuracy under knowledge conflicts. In summary, the main contributions of this work are as follows: 
\begin{itemize}
    \item We propose \textbf{CC-VQA}, a training-free framework that mitigates knowledge conflicts in KB-VQA through vision-centric contextual reasoning and correlation-guided generation. The approach externalizes parametric knowledge for conflict analysis, prioritizing core conflicts during answer synthesis.
    \item We introduce correlation-aware positional compression for low-correlation content and adaptive decoding with correlation-weighted conflict scoring. These improve conflict resolution while reducing noise sensitivity.
    \item Extensive experiments demonstrate our method's efficiency and effectiveness, achieving new state-of-the-art results on E-VQA, InfoSeek and OK-VQA benchmarks, outperforming complex alternatives with higher efficiency.
\end{itemize}

%% file: sec/2_related_work.tex
\section{Related Work}
\label{sec:related}

\begin{figure}[t]
    \centering
    \includegraphics[width=1\linewidth]{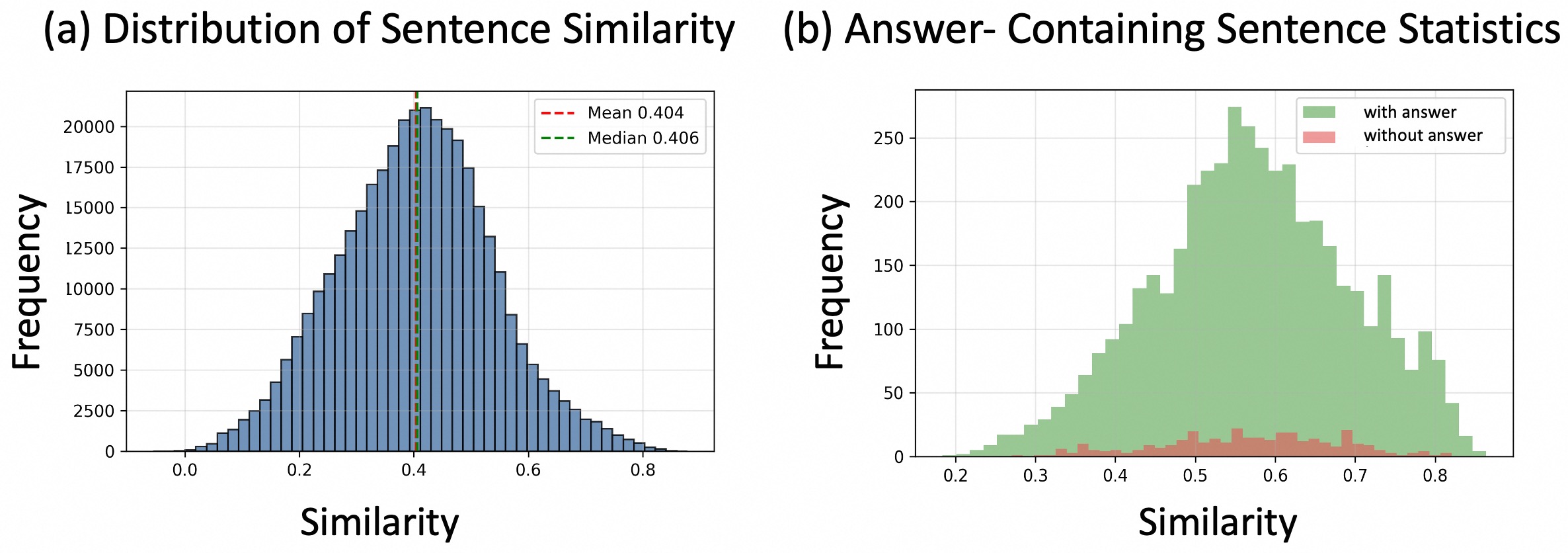}
    \caption{\textbf{Similarity Statistics Between Contextual Sentences and Question.}}
    \label{fig:context_similarity_stat}
    \vspace{-0.3cm}
\end{figure}

\subsection{Knowledge-Based Visual Question Answering}
Knowledge Based Visual Question Answering (KB-VQA)~\cite{deng2025comprehensive}, as a key branch of visual question answering (VQA)~\cite{antol2015vqa,goyal2017making}, enhances model performance by guiding it to retrieve entity knowledge to reason and answer visual questions. Existing KB-VQA methods follow a three-stage pipeline consisting of retrieval, reranking, and generation~\cite{qi2024rora, tian-etal-2025-core,ling2025mmkb,yan2024echosight,caffagni2024wiki,cocchi2025augmenting, yang2025omgm}. In retrieval, current methods~\cite{yan2024echosight,yang2025omgm} primarily focus on achieving fine-grained retrieval, Wiki-LLaVA enhances the model’s knowledge reasoning by incorporating external multimodal documents as context through hierarchical retrieval. Recently, EchoSight~\cite{yan2024echosight} retrieves target sections based on image-to-image similarity and then re-ranks them using a trained reranker model. In generation, current RAG methods~\cite{cocchi2025augmenting,hongknowledge} primarily aim to process retrieved information to reduce redundancy. ReflectiVA~\cite{cocchi2025augmenting} uses reflection tokens to let the model dynamically assess and manage external knowledge via two-stage training, preserving performance when external knowledge is unnecessary. Wiki-PRF~\cite{hongknowledge} trains the model via reinforcement learning to filter retrieved information and select the most useful content. In this paper, we guide the model to leverage reasons and similarity scores to handle redundant retrieved information.

\begin{figure*}[t]
    \centering
    \includegraphics[width=1\linewidth]{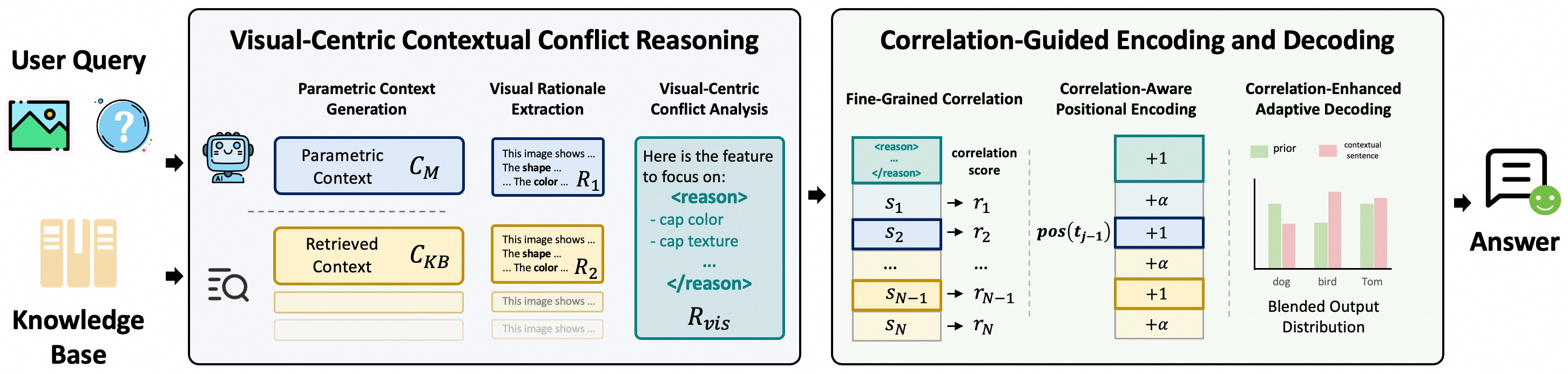}
    \caption{\textbf{Overview of CC-VQA.} CC-VQA consists of two key components.(1) Visual-Centric Contextual Conflict Reasoning: CC-VQA extracts semantic visual descriptions from both parametric and external contexts, then reasons about knowledge conflicts with the summarized visual-centric information.(2) Correlation-Guided Encoding and Decoding: By computing fine-grained correlations, CC-VQA dynamically compresses low-correlation information during positional encoding and adjusts the output distribution during decoding based on these correlations.} 
    \label{fig:overview}
\end{figure*}

\subsection{Knowledge Conflict in KB-QA}
Knowledge conflicts~\cite{mingfaitheval,ying2024intuitive,yuan2024discerning,longpre2021entity} in KB-QA refers to the discrepancies between retrieved external knowledge and the model’s internal parametric knowledge. In early work, Xie et al. \citep{xie2023adaptive} propose a framework to probe LLMs’ parametric and counter-memory, revealing strong confirmation bias, sensitivity to evidence presentation, and risks of generating convincing misinformation. Recently, FaithfulRAG~\cite{zhang2025faithfulrag} identifies factual conflicts and uses a self-reflection process to reconcile them before response generation. Compared to these methods~\cite{zhou2023context,zhang2023merging} that rely on prompt engineering or training auxiliary discriminators to handle knowledge conflicts in RAG, constrastive decoding methods~\cite{shi2024trusting, li2015diversity, liu2021dexperts, zhao2024enhancing, wang2025adacad, khandelwal2025cocoa} have garnered attention due to their training-free nature. AdaCAD~\cite{wang2025adacad} dynamically adjusts decoding based on Jensen-Shannon divergence between contextual and parametric distributions, improving QA accuracy and summary factuality while remaining robust in both conflicting and non-conflicting cases. CoCoA~\cite{khandelwal2025cocoa} employs Rényi divergence and entropy-based confidence measures in its decoding process to adaptively resolve knowledge conflicts and enhance generation faithfulness. However, existing methods do not account for the relevance between retrieved information and the query; in our approach, we explicitly incorporate sentence relevance into knowledge conflict.


%% file: sec/3_observation.tex
\section{Problem and Observation}
\label{sec:observation}

\textbf{Knowledge-based Visual Question Answering (KB-VQA).} 
Knowledge-based Visual Question Answering requires generating an answer $A$ to a question $Q$ about an input image $I$ by utilizing external knowledge from a structured knowledge base $\mathcal{KB} = \{(a_1, I_1), \dots, (a_n, I_n)\}$, where each pair consists of an entity article $a_i$ and its associated image $I_i$. In multimodal Retrieval-Augmented Generation (RAG) frameworks, given the query $(I, Q)$, the system retrieves relevant knowledge entries $\{(a_j, I_j)\}$ through a multimodal similarity function. From each retrieved article $a_j$, relevant text sections are selected to construct a knowledge context $C$. The final answer $A$ is generated by conditioning a vision language model on the input image $I$, question $Q$, and aggregated context $C$.

\textbf{Knowledge Conflict in KB-VQA.}
In retrieval-augmented generation (RAG) based multimodal KB-VQA systems, conflicts between parametric knowledge $\Theta$ embedded in vision language models and externally retrieved knowledge context $\mathcal{C}$ from $\mathcal{KB}$ degrade answer quality, leading to inaccuracies or internal inconsistencies. We analyzed 10K samples from the InfoSeek benchmark using the Qwen2.5-VL-7B multimodal RAG framework. While this approach achieved a $16.82\%$ accuracy gain over direct VLM answering, it introduced errors in $10.53\%$ of cases previously answered correctly by the VLM alone. This performance degradation stems from knowledge conflicts, as illustrated in Figure~\ref{fig:knowledge_conflict}. Our goal is therefore to mitigate knowledge conflicts between retrieval contexts and parametric knowledge, thereby improving KB-VQA answer accuracy.

\textbf{Observation 1.}
\emph{Visual semantic features are helpful for identifying contextual conflicts. }
As illustrated in Figure~\ref{fig:knowledge_conflict}, similarity-based retrieval using image-text features in RAG frameworks may yield inaccurate contexts that conflict with parametric knowledge, potentially misleading answer generation. 
However, retrieved contexts, regardless of their accuracy, often contain textual descriptions of visual attributes (e.g., spatial relationships, colors, shapes) found in the query image $I$. Consequently, the visual information in $I$ can be leveraged to validate these textual claims and resolve conflicts.

\textbf{Observation 2.}
\emph{Retrieved contexts frequently contain substantial redundancy, making fine-grained analysis necessary. }
We analyzed 10K random samples from the InfoSeek benchmark, employing BLIP to compute sentence-level similarity between retrieved contexts and query pairs $(Q, I)$. Figure~\ref{fig:context_similarity_stat} presents the similarity distribution: (a) reveals normally distributed similarities (mean $\mu=0.4$, $\sigma=0.15$) with fewer than $0.3\%$ of sentences exceeding similarity 0.8. Subfigure (b) analyzes the top-3 retrieved contexts per question, revealing that each context contains an average of 107 sentences. When ranking sentences by similarity, the correct answers for $90\%$ of questions reside within the top $25\%$ highest-similarity sentences.

%% file: sec/4_method.tex
\section{Method}
\label{sec:method}

\subsection{Overview}
To address knowledge conflicts between model-parametric and external knowledge in KB-VQA, we propose \textbf{C}onflict- and \textbf{C}orrelation-aware \textbf{VQA} (CC-VQA), which focuses on the generation stage. As depicted in Figure~\ref{fig:overview}, the method first performs \emph{Visual-Centric Contextual Conflict Reasoning} to externalize parametric knowledge and explicitly infer conflicts across all contextual sources. These analyzed contexts then drive a \emph{Correlation-Guided Encoding and Decoding process}, where correlation-aware positional encoding compression focuses on query-salient statements during encoding, while adaptive decoding adjusts token generation based on key correlations for conflict resolution.
By characterizing context-level conflicts and dynamically prioritizing correlated statements during generation, our approach mitigates answer inaccuracies arising from knowledge inconsistencies.

\subsection{Preliminary Study}
Unlike additive position encodings, Rotary Position Embedding (RoPE)~\cite{su2024roformer} encodes the absolute position $m$ as a rotation operator applied to the embedding vector. Specifically, for an input embedding $\mathbf{x} \in \mathbb{R}^{d}$ at position $m$, the RoPE-transformed representation is given by:
\begin{equation}
\text{RoPE}(\mathbf{x}, m) = \mathbf{R}_m \mathbf{x}, 
\end{equation}
\begin{equation}
[\mathbf{R}_m]_{2i-1:2i} = 
\begin{bmatrix}
\cos(m\theta_i) & -\sin(m\theta_i) \\
\sin(m\theta_i) & \cos(m\theta_i)
\end{bmatrix},
\end{equation}
where $\theta_i = 10000^{-2i/d}$, for $i = 1, \dots, d/2$. Building upon RoPE, Position Interpolation (PI)~\cite{chen2023extending,pengyarn} extends the context length beyond the pre-training limit by scaling input positions $m$ to $m / \alpha$ (where $\alpha > 1$). This ensures that all effective positions remain within the original pre-trained range, enabling effective fine-tuning on longer sequences.

\subsection{Visual-Centric Contextual Conflict Reasoning}
Resolving knowledge conflicts fundamentally requires identifying and analyzing inconsistencies between retrieved external knowledge and model-parametric knowledge. Unlike conventional conflict detection approaches, our method performs vision-centric conflict analysis at the contextual level for KB-VQA, examining how knowledge discrepancies correlate with visual semantic features in queries. We first generate parametric knowledge contexts conditioned on the user query. Subsequently, through visual analysis, we extract knowledge-vision associations across contexts with the query image and question. Based on these associations, we abstract and generalize visual semantic patterns to infer conflict-relevant visual features. This vision-oriented conflict characterization provides critical guidance for conflict resolution during answer generation.

\textbf{Parametric Context Generation.}
Unlike retrieved contexts $\{C_{KB}\}$ from knowledge bases, we employ VLM to generate parametric contexts analogous to $C_{KB}$. Specifically, given a user query comprising an image $I$ and question $Q$, we prompt the VLM to generate both the answer $A$ and relevant background knowledge, constituting the parametric context $C_M$. This $C_M$ represents a concrete externalization of model-parametric knowledge, encompassing supporting evidence and explicit answer while maintaining structural consistency with retrieved contexts. We then combine $C_M$ and $C_{KB}$ into a context set $\mathcal{C}=\{C_M,C_{KB}\}$ for context-level knowledge conflict reasoning.

\textbf{Visual Rationale Extraction.}
Based on the context corpus $\mathcal{C}$, we leverage the VLM to analyze logical relationships between each context $C$ and the query image $I$. Specifically, we identify which visual semantic features correlate with contextual knowledge conclusions. This process is formalized as:
\begin{equation}
    R_i = \text{VLM}(I, Q, C_i), \quad \forall C_i \in \mathcal{C},
\end{equation}
where $R_i$ denotes the visual reasoning output for context $C_i$. The objective is to clarify image features critically relevant to contextual knowledge, enabling the model to prioritize these visual semantics during answer generation and mitigate knowledge conflicts arising from visual misinterpretation.

\begin{figure}[t]
    \centering
    \includegraphics[width=1\linewidth]{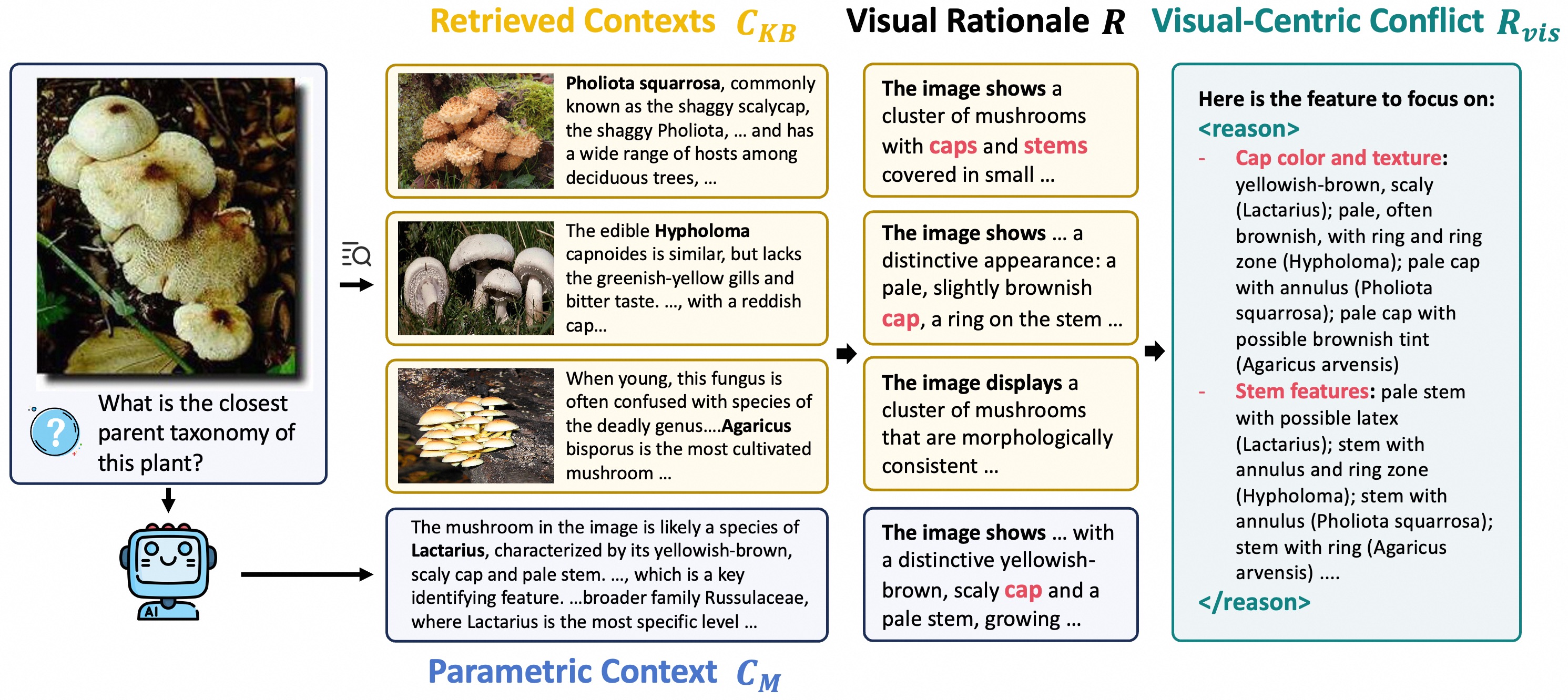}
    \caption{\textbf{Illustration of Visual-Centric Contextual Conflict Reasoning}. CC-VQA explicitly extracts the model’s parametric context and the visual rationale for all contexts. Through visual rationales, CC-VQA identifies and summarizes visual-centric conflicts $R_{vis}$ to address conflicting information.}
    \label{fig:visual_centric}
    \vspace{-0.5cm}
\end{figure}

\textbf{Visual-Centric Conflict Analysis.}
Based on visual reasoning descriptions $\{R_i\}$ derived from all contexts, we abstract semantic features characterizing knowledge conflicts, for example, divergent answers originating from distinct stem characteristics in mushrooms (Figure~\ref{fig:visual_centric}). This process adaptively summarizes the core conflict points and critical visual details, providing explicit visual cues for answer generation. Formally, we employ prompt-guided VLM summarization:  
\begin{equation}
    R_{vis} = \text{VLM}\left(I, Q, \{R_i\}\right),
\end{equation}  
where $R_{vis}$ encapsulates key visual conflict information within structured reasoning tags \textcolor{teal}{$<$reason$>$}\textcolor{teal}{$<$/reason$>$}. Figure~\ref{fig:visual_centric} illustrates an example.

\subsection{Correlation-Guided Encoding and Decoding}

In the preceding stage, we externalized model-parametric knowledge and constructed a contextual corpus $\mathcal{C}$ by integrating retrieved external knowledge. This corpus serves as the primary source for final answer generation. However, each context $C_i \in \mathcal{C}$ contains both answer-relevant statements and extraneous content. Such irrelevant information acts as noise that may impair the model's ability to resolve knowledge conflicts. To address this limitation, we propose a fine-grained Correlation-Guided Encoding-Decoding approach. 
Specifically, our method conducts relevance assessment at the statement level across all $C_i$ contexts relative to the user question $Q$. During encoding, we apply positional encoding compression to attenuate low-relevance statements, focusing attention on critical conflict regions. During decoding, we enhance adaptive token sampling using statement relevance weights to improve answer accuracy under knowledge conflict scenarios.

\textbf{Fine-Grained Correlation.}
For each context $C_i$, we conduct fine-grained relevance analysis with the user question by decomposing $C_i$ into constituent statements, i.e., sentences. This enables identification of sentences containing information related to the question, critical supporting details, or key sources of knowledge conflicts that require prioritized attention during generation. Formally, given a context $C_i$ comprising $N$ sentences $\{s_{ij}\}_{j=1}^N$, we employ EVA-CLIP to estimate the relevance score $r_{ij}$ between each sentence $s_{ij}$ and image-question pair $(I,Q)$.  
To mitigate abstraction and ambiguity in the original user's question $Q$, we first rewrite $Q$ using an image-grounded prompt that explicitly resolves entity references and visual features, yielding a disambiguated question $Q^*$. The sentence-level correlation is then computed as:  
\begin{equation}
    r_{ij} = \frac{1}{2}\left({\text{EVA-CLIP}(Q^*, s_{ij}) + \text{EVA-CLIP}(I, s_{ij})}\right).
\end{equation}
Consequently, we associate every sentence across all contexts with a relevance metric, formalized as $C_i^{*} = \{(s_{ij}, r_{ij}) \mid j=1,\ldots,N\}$.

\textbf{Correlation-Aware Positional Encoding.}
During generation, we integrate $R_{vis}$ and $\mathcal{C}$ with the user query (image $I$ and question $Q$) as contextual inputs. $\mathcal{C^{*}}$  is provided as supplementary information for later use. To enhance focus on high-relevance statements, we modify positional encoding based on empirical findings: accurate answers predominantly reside in highly relevant contextual statements, whereas low-relevance statements often contain transitional content with minimal core information. Instead of amplifying attention to critical statements, our method compresses the relative length of low-relevance statements, reducing their attentional allocation and prioritizing core conflict resolution.

Specifically, positional encodings are constructed sequentially but dynamically adjusted according to statement-level relevance. First, we sort all sentence pairs $\{(s_k, r_k)\}$ by descending $r_k$, defining the low-correlation set as the bottom $\tau$ percentile:
\begin{equation}
\mathcal{L}_{\tau} = \{ s_{(N-b+1)}, s_{(N-b+2)}, \dots, s_{(N)} \},
\end{equation}
where $\quad b = \lfloor \tau \times N \rfloor$, and $s_{(k)}$ denotes the $k$-th highest correlation sentence. Let $\operatorname{sent}(t_j)$ indicate the sentence containing token $t_j$. The position index updates as:
\begin{equation}
\operatorname{pos}(t_j) = 
\begin{cases} 
\operatorname{pos}(t_{j-1}) + \alpha & \text{if } \operatorname{sent}(t_j) \in \mathcal{L}_{\tau}, \\ 
\operatorname{pos}(t_{j-1}) + 1 & \text{otherwise},
\end{cases}
\end{equation}
where $\alpha=0.5$ scales position increments for low-correlation tokens. Since $R_{vis}$ provides visual conflict analysis independent of positional order, its tokens retain original positional encodings. 
By preserving full positional resolution for high-correlation sentences while compressing the positional space of low-correlation sentences, our method maintains focused attention on semantically critical content and delivers efficient position-aware representations for extended sequences. 

\textbf{Correlation-Enhanced Adaptive Decoding.}
During decoding, we enhance the influence of high-correlation contextual sentences on token sampling distributions. This design stems from the observation that such sentences indicate higher knowledge conflict probability and thus exert stronger impact. Our correlation-enhanced adaptive decoding, driven by fine-grained contextual correlation, outperforms section-level adjustments by precisely targeting influential sentences, as not all contextual content affects the answer.  

Building upon adaptive contrastive decoding~\cite{khandelwal2025cocoa}, we augment conflict scoring with fine-grained correlation beyond distribution divergence ($D_t$) and entropy gap ($\Delta H_t$). The enhanced conflict score is:  
\begin{equation}
s'_t = \sigma ( D_t + \Delta H_t + K  + \delta ),
\end{equation}  
where $\sigma$ denotes the sigmoid function, and $K$ combines average correlation and its concentration, with $\delta$ being a small bias term (e.g., $\delta=0.1$).
\begin{equation}
    K = 1 - \left(\frac{1}{N}\sum_{i=1}^N r_i\right) \cdot \left(1 - \frac{H(r)}{\log M}\right),
\end{equation}
where $H(\mathbf{r}) = -\sum_{i=1}^M r_i \log r_i$. The first part measures average sentence correlation, while the second part quantifies correlation concentration (distinguishing focused strong evidence from dispersed weak evidence). 

Through this procedure, samples exhibiting high divergence, large entropy gap, and low/dispersed correlation receive elevated conflict scores, ultimately modulating the blended output distribution.

\begin{table*}[t]
  \centering
  \setlength{\tabcolsep}{.4em}
  \resizebox{\linewidth}{!}{
  \begin{tabular}{lc c c @{\hspace{0em}} cc c ccc}
   \toprule
    & & & & \multicolumn{2}{c}{\textbf{E-VQA}} & & \multicolumn{3}{c}{\textbf{InfoSeek}} \\
    \cmidrule{5-6} \cmidrule{8-10}
     \textbf{Method} & \textbf{Model} & \textbf{Gen. FT} & \textbf{Retriever} & Single-Hop & All & & Unseen-Q & Unseen-E & All \\
    \midrule
    \rowcolor{black!15} 
    \multicolumn{10}{l}{\textit{Zero-shot MLLMs}} \\
    BLIP-2~\cite{li2023blip} & Flan-T5$_\text{XL}$ & – & – & 12.6 & 12.4 & & 12.7 & 12.3 & 12.5 \\
    InstructBLIP~\cite{dai2023instructblip} & Flan-T5$_\text{XL}$ & – & – & 11.9 & 12.0 & & 8.9 & 7.4 & 8.1 \\
    LLaVA-v1.5~\cite{liu2024improved} & Vicuna-7B & – & – & 16.3 & 16.9 & & 9.6 & 9.4 & 9.5 \\
    GPT-4V~\cite{achiam2023gpt} & – & – & – & 26.9 & 28.1 & & 15.0 & 14.3 & 14.6 \\
    Qwen2.5-VL-7B~\cite{bai2025qwen2} & – & – & – & 21.7 & 20.3 & & 22.8 & 24.1 & 23.7 \\
    \midrule
    \rowcolor{black!15}  
    \multicolumn{10}{l}{\textit{Retrieval-Augmented Models}} \\
    DPR$_\text{V+T}$~\cite{lerner2024cross}$^\dagger$ & Multi-passage BERT & – & CLIP ViT-B/32 & 29.1 & – & & – & – & 12.4 \\
    RORA-VLM~\cite{qi2024rora}$^\dagger$ & Vicuna-7B & \ding{52} & CLIP+Google Search & – & 20.3 & & 25.1 & 27.3 & – \\
    EchoSight~\cite{yan2024echosight}$^\dagger$ & Mistral-7B/LLaMA-3-8B & \ding{55} & EVA-CLIP-8B & 26.4 & 24.9 & & 30.0 & 30.7 & 30.4 \\
    Wiki-LLaVA~\cite{caffagni2024wiki} & Vicuna-7B & \ding{52} & CLIP ViT-L/14+Contriever & 17.7 & 20.3 & & 30.1 & 27.8 & 28.9 \\
    ReflectiVA~\cite{cocchi2025augmenting} & LLaMA-3.1-8B & \ding{52} & EVA-CLIP-8B & 28.0 & 29.2 & & 40.4 & 39.8 & 40.1 \\
    MMKB-RAG~\cite{ling2025mmkb} & Qwen2-7B & \ding{55} & EVA-CLIP-8B & 39.7 & 35.9 & & 36.4 & 36.3 & 36.4 \\
    Wiki-PRF~\cite{hongknowledge} & Qwen2.5-VL-8B & \ding{52} & EVA-CLIP-8B & 37.1 & 36.0 & & 43.3 & 42.7 & 42.8 \\
    Qwen2.5-VL-7B~\cite{bai2025qwen2} & Qwen2.5-VL-7B & \ding{55} & EVA-CLIP-8B$^{*}$ & 36.7 & 31.2 & & 41.9 & 41.3 & 41.8 \\
    \rowcolor{cyan!15}
    CC-VQA (\textbf{Ours}) & Qwen2.5-VL-7B & \ding{55} & EVA-CLIP-8B$^{*}$ & \textbf{41.4} & \textbf{36.1} & & \textbf{44.7} & 
    \textbf{46.1} & \textbf{45.1} \\
  \bottomrule
  \end{tabular}
  }
  \caption{\textbf{VQA accuracy on E-VQA and InfoSeek.} Our method is highlighted in light blue. The superscript $^{*}$ denotes the use of the top-3 sections to intentionally induce knowledge conflict, and Gen.FT indicates whether the generation process is fine-tuned.}
  \label{tab:results}
  \vspace{-0.35cm}
\end{table*}

%% file: sec/5_experiments.tex
\section{Experiments}
\subsection{Datasets}
We evaluate our method on three benchmark datasets: 
(1) \textbf{Encyclopedic VQA (E-VQA)}~\cite{lin2023fine} contains 221K+ distinct QA pairs, each associated with up to five images from iNaturalist~\cite{van2021benchmarking} and Google Landmarks v2~\cite{weyand2020google}. Questions are categorized as single-hop or two-hop, with dataset splits of 1M training, 13.6K validation, and 5.8K test instances. 
(2) \textbf{InfoSeek}~\cite{lin2022retrieval} comprises 1.3M VQA pairs grounded in 11K OVEN images~\cite{hu2023open}. Its training (934K) and validation (73K) sets maintain entity/question disjointness, with the validation set partitioned into Unseen Entity and Unseen Question subsets. Following~\cite{yan2024echosight}, we utilize a 100K-article Wikipedia knowledge base and evaluate on the full validation set.
(3) \textbf{OK-VQA}~\cite{marino2019ok}, a KB-VQA benchmark built on COCO, contains 14K questions. We employ the InfoSeek knowledge base for experiments on this dataset.

\subsection{Metrics}
In the KB-VQA task, as our primary focus lies in the generation process, we rely on question-answering metrics to evaluate performance. Specifically, following the original dataset protocols, we use VQA accuracy~\cite{goyal2017making,marino2019ok} for InfoSeek and BEM score\cite{zhangbertscore} for E-VQA.

\subsection{Implementation Details}
We implement our method using the publicly available Qwen2.5-VL-7B model. For retrieval, we employ a frozen EVA-CLIP-8B model, utilizing Echosight's reranker~\cite{yan2024echosight} for both retrieval and reranking stages. 
The retrieval pipeline first indexes and retrieves the top-20 articles using cosine similarity on image features via the Faiss-GPU library, followed by selection of the top-3 relevant sections. Our approach is entirely training-free. 
During inference on an 8×A800 GPU system, our method completes the full evaluation in 8 hours.


\begin{figure*}[!t]
    \centering
    \includegraphics[width=1\linewidth]{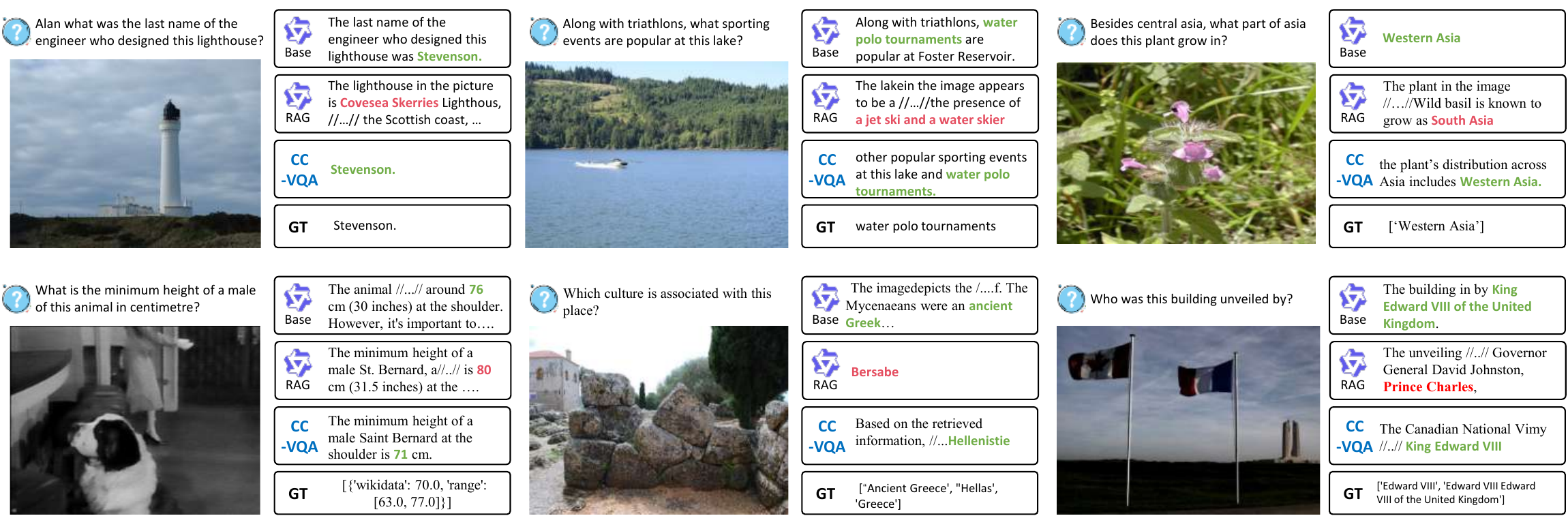}
    \caption{\textbf{The Case of CC-VQA}. Qualitative results from Encyclopedic-VQA (top row) and InfoSeek (bottom row) illustrate knowledge conflicts introduced by retrieved information and show the effective mitigation achieved by our model.} 
    \label{fig:case_cc-vqa}
    \vspace{-0.3cm}
\end{figure*}

\subsection{Main Results}
\textbf{Results on E-VQA and Infoseek.}
Table~\ref{tab:results} presents experimental results for E-VQA and InfoSeek datasets. We report performance for both zero-shot multimodal large language models (MLLMs) without retrieval and their retrieval-augmented counterparts. To isolate the impact of knowledge conflict mitigation, we explicitly indicate whether each model's generator underwent fine-tuning. 
Standard retrieval augmentation establishes strong baselines, boosting Qwen2.5-VL-7B performance by $+11.2\%$ on E-VQA and $+18.1\%$ on InfoSeek. Our method achieves further improvements of $+4.7\%$ and $+3.3\%$ respectively, demonstrating effective knowledge conflict resolution. Notably, our training-free approach outperforms reinforcement learning-based Wiki-PRF and significantly exceeds MMKB-RAG (another fine-tuning-free method) by $+5.1\%$ on InfoSeek.

\noindent \textbf{Results on OK-VQA.}
Our method achieves state-of-the-art performance on OK-VQA with a $78.8\%$ accuracy as shown in Table~\ref{tab:okvqa}. This training-free result surpasses both existing non-fine-tuned approaches and the reinforcement learning-based Wiki-PRF method~\cite{hongknowledge}, demonstrating superior knowledge conflict resolution capabilities.

\begin{table}[!t]
\centering
\caption{\textbf{Performance on OK-VQA.}}
\label{tab:okvqa}
\resizebox{0.45\textwidth}{!}{
\begin{tabular}{lc cc cc cc}
\toprule
Method &Model & Gen.FT & Accuracy\\ \midrule
Qwen2.5-VL-7B &-& - &72.4 \\
Wiki-PRF-7B~\cite{hongknowledge} & Qwen2.5-VL-7B&\ding{52} & 77.8 \\
MMKB-RAG~\cite{ling2025mmkb}& LLaMA-3.1-8B&\ding{55} & 65.4 \\
KU-RAG~\cite{zhang2025fine} &LLaVA-Next-7B&\ding{55} & 73.1 \\
CC-VQA (\textbf{Ours}) & Qwen2.5-VL-7B &\ding{55} &78.8\\
\bottomrule
\end{tabular}}
\end{table}

\noindent \textbf{Oracle Analysis.}
We establish an upper-bound performance estimate through oracle experiments on 10K InfoSeek samples, see Table~\ref{tab:oracles}. By manually inserting ground-truth sections into the top-3 retrieved contexts, we simulate near-perfect retrieval conditions. Our method achieves significantly higher VQA accuracy (66.5\%) than baselines, demonstrating superior utilization of oracle information for efficient knowledge localization.

\begin{table}[!t]
\centering
\caption{\textbf{Oracle Analysis.} VQA Accuracy in Oracle Setting with Ground-Truth Articles.}
\label{tab:oracles}
\resizebox{0.45\textwidth}{!}{
\begin{tabular}{lc cc cc cc}
\toprule
Method & Model & Retriever & Accuracy \\ \midrule
Qwen2.5-VL-7B & - & EVA-CLIP-8B$^{*}$ & 55.3\\
Ours & Qwen2.5-VL-7B &EVA-CLIP-8B$^{*}$ & 66.5\\
\bottomrule
\end{tabular}}
\end{table}

\noindent \textbf{Benefits of Knowledge Conflict Mitigation.}
To more intuitively demonstrate the benefits of mitigating knowledge conflict, we conduct the following analysis on 10K InfoSeek samples. Specifically, we first asked the model to directly answer user queries, and then used the Vanilla RAG method to answer the same questions. We separately calculated the proportion of newly added correct answers and the proportion of newly added incorrect answers after applying the RAG method, defining them as the Helpful Ratio and Harmful Ratio of the RAG method, respectively. Similarly, we repeated the above experiment using the CC-VQA method. Table~\ref{tab:Knowledge Conflict Ratio} shows the results and changes in the Helpful Ratio and Harmful Ratio. Our method significantly reduced the Harmful Ratio from $10.53\%$ to $7.69\%$, while also increasing the Helpful Ratio from $16.82\%$ to $18.63\%$, demonstrating the effectiveness of our method in reconciling the discrepancies between model parametric knowledge and external knowledge.

\begin{table}[!t]
\centering
\caption{\textbf{Benefits of Knowledge Conflict Mitigation.} }
\label{tab:Knowledge Conflict Ratio}
\resizebox{0.45\textwidth}{!}{
\begin{tabular}{lc cc cc cc}
\toprule
Method & Retriever & Helpful Ratio & Harmful Ratio \\ \midrule
Qwen2.5-VL-7B & EVA-CLIP-8B$^{*}$ &16.82  & 10.53\\
CC-VQA(\textbf{Ours}) & EVA-CLIP-8B$^{*}$ &18.63 & 7.69 \\
\bottomrule
\end{tabular}}
\end{table}

\subsection{Ablation Study}

\noindent \textbf{Component Ablation.}
Table~\ref{tab:Component Ablation} presents ablation studies on 10K InfoSeek subsamples for our core modules: Visual-Centric Contextual Conflict Reasoning (VCCR), Correlation-Aware Positional Encoding (CPE), and Correlation-Enhanced Adaptive Decoding (CAD). Introducing \emph{VCCR} yields a $1.9\%$ accuracy gain over Vanilla RAG, demonstrating that vision-centric conflict identification enhances conflict resolution. Adding \emph{CAD} further improves accuracy by $0.8\%$ by adapting output distributions during decoding. Incorporating \emph{CPE} provides an additional $0.9\%$ gain through increased focus on high-correlation context sentences. These experiments validate the efficacy of each CC-VQA component.

\begin{table}[!t]
\centering
\caption{\textbf{Component Ablation.} We perform ablation studies on the components of CC-VQA using a 10K subset of InfoSeek.}
\label{tab:Component Ablation}
\resizebox{0.45\textwidth}{!}{
\begin{tabular}{lc cc cc cc}
\toprule
Model & Retriever & VCCR & CAD & CPE & Accuracy \\ \midrule
Qwen2.5-VL-7B & EVA-CLIP-8B$^{*}$ & & &  &41.4   \\
Qwen2.5-VL-7B& EVA-CLIP-8B$^{*}$&\ding{52} & &   &43.3  \\
Qwen2.5-VL-7B & EVA-CLIP-8B$^{*}$&\ding{52}  &\ding{52} &   &44.1 \\
Qwen2.5-VL-7B & EVA-CLIP-8B$^{*}$&\ding{52}  &\ding{52} & \ding{52}  &45.0 \\
\bottomrule
\end{tabular}}
\end{table}

\noindent \textbf{Ablation of Alpha.}
Table~\ref{tab:Albation of a} presents the ablation study for the compression parameter $\alpha$ in positional encoding, evaluated from 0.1 to 1.0. Results show accuracy gradually decreases as $\alpha$ decreases due to excessive semantic compression. Even at $\alpha=0.1$, the method maintains high accuracy, supporting our observation that contexts contain substantial redundant information irrelevant to user queries. Compressing such contextual sentences minimally affects response accuracy, while appropriate $\alpha$ selection (set to 0.5) maximizes utilization of highly relevant sentences, thereby improving answer accuracy.

\begin{table}[!t]
\centering
\caption{\textbf{Ablation of Alpha.} We conduct ablation studies varying the encoding compression parameter ($\alpha$) from $0.1$ to $1.0$.}
\label{tab:Albation of a}
\begin{tabular}{lc cc cc cc}
\toprule
$\alpha$ &0.1 & 0.3 & 0.5 & 0.7& 0.9 & 1.0 \\ \midrule
Accuracy &44.0  &44.2 &45.0 &44.9  &44.5 & 44.1\\
\bottomrule
\end{tabular}
\end{table}

\begin{figure}[h]
    \centering
    \includegraphics[width=1\linewidth]{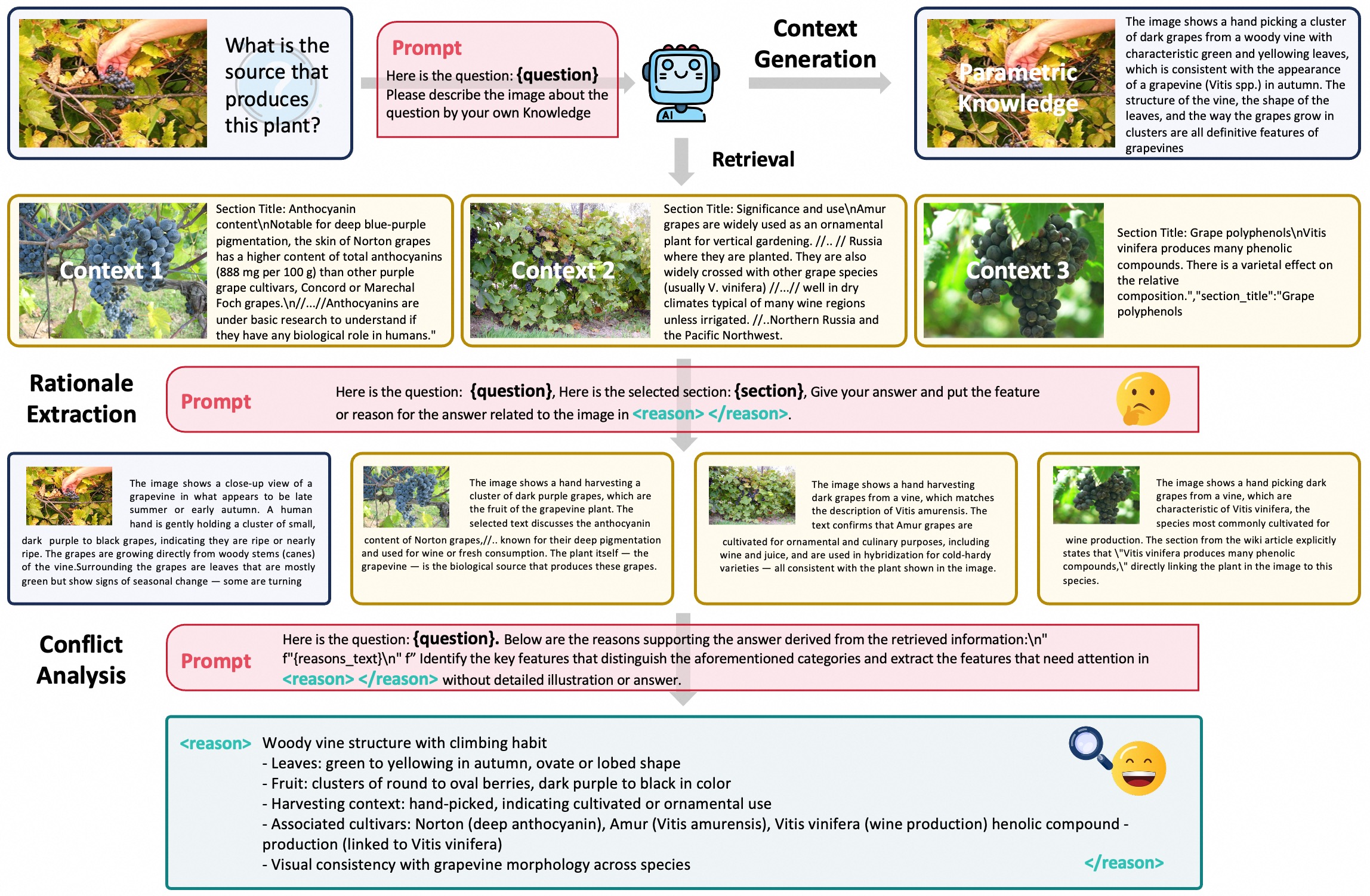}
    \caption{\textbf{Illustration of Visual-Centric Contextual Conflict Reasoning}. The figure shows how CC-VQA extracts visual-centric conflicts, with the corresponding image regions highlighted by red boxes.}
    \label{fig: Case of VCCR}
\end{figure}

\subsection{Case Studies}
\noindent \textbf{The Case of CC-VQA.}
We demonstrate our method's efficacy using six representative samples from E-VQA and InfoSeek spanning architecture, sports, plants, animals, culture, and country (Figure~\ref{fig:case_cc-vqa}). Specifically, ``Base'' indicates the model without retrieval (parametric knowledge only), ``Retrieval'' incorporates external knowledge, and ``CC-VQA'' denotes our approach. Analysis shows external knowledge introduces beneficial information but may conflict with and override correct parametric knowledge in the model. Our method resolves such conflicts through visual-centric conflict analysis and fine-grained correlation-guided generation, consistently preserving accurate answers.

\noindent \textbf{The Case of Visual-Centric Contextual Conflict Reasoning.} 
Figure~\ref{fig: Case of VCCR} illustrates the visual-centric contextual conflict reasoning process. For the question about the plant's parent taxonomy, the model first externalizes its internal knowledge and synthesizes it with retrieved external facts. 
This synthesis produces a summary of key characteristics. Crucially, this summary then acts as an explicit prompt, guiding the model to focus on the relevant visual details in the image (e.g., leaf shape, flower structure) to correctly answer the question. 
This step-by-step reasoning compels the model to ground its final answer in visual evidence.

\begin{figure}[!t]
    \centering
    \includegraphics[width=1\linewidth]{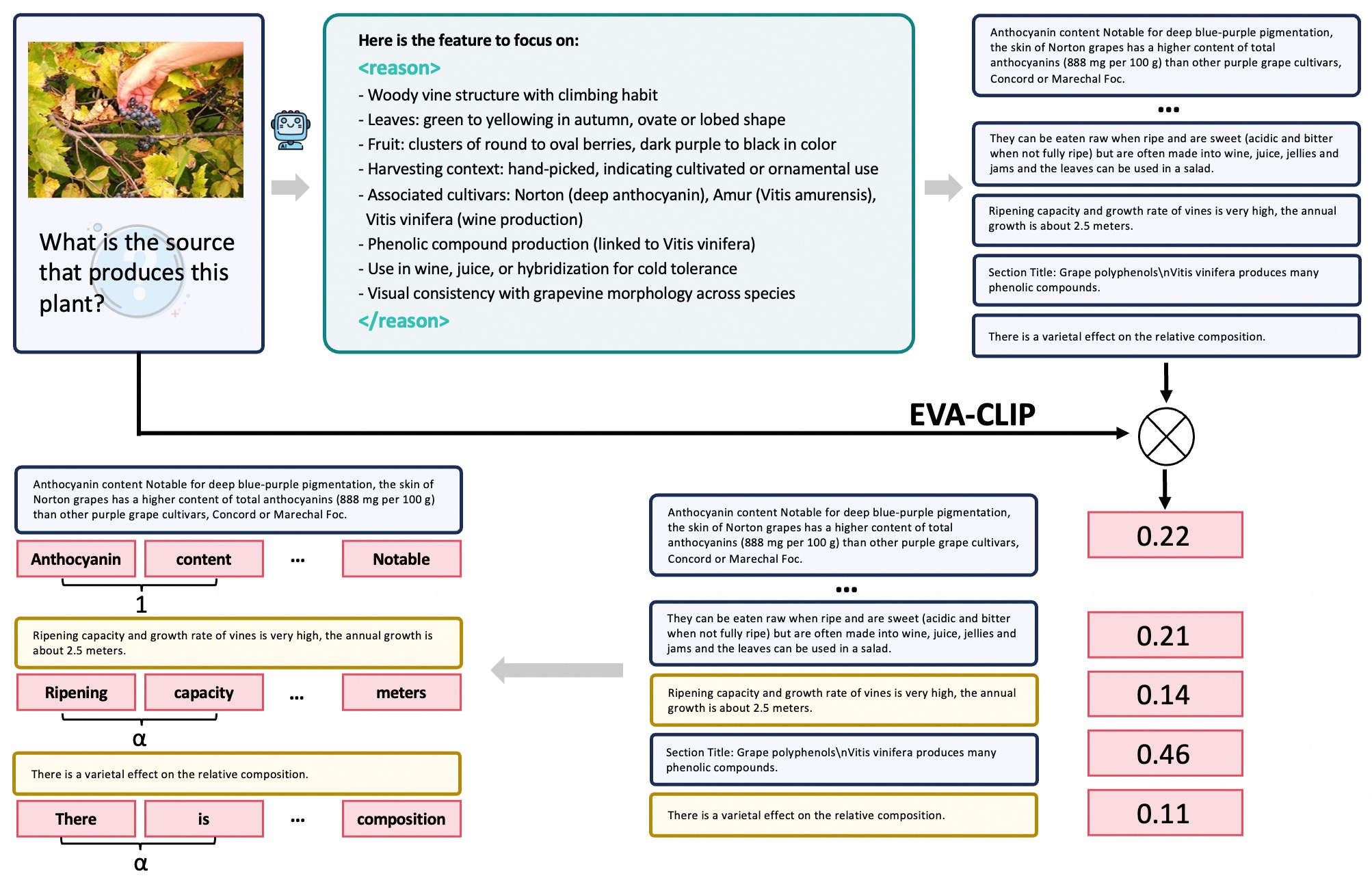}
    \caption{\textbf{Illustration of Correlation-Aware Positional Encoding.} }
    \label{fig: Case of CPE}
\end{figure}

\noindent \textbf{The Case of Correlation-Aware Positional Encoding.}
Figure~\ref{fig: Case of CPE} provides a visual demonstration of our correlation-aware positional encoding module in action. 
To illustrate its effect, we represent sentence similarity scores through font size, which decreases as similarity drops from a high of 0.48 to a low of 0.1.
Critically, the figure shows that the sentence with the highest similarity score (0.48) is precisely the one containing the ground-truth answer, ``Amanita," demonstrating the module's ability to extract crucial information.

%% file: sec/6_conclusion.tex
\section{Conclusion}
\label{sec:conclusion}

In this work, we address knowledge conflicts in knowledge-based visual question answering through two key observations, leading to the proposed CC-VQA method, a conflict- and correlation-aware solution for mitigating knowledge conflict in KB-VQA. CC-VQA integrates two core components: (1) \emph{Visual-Centric Contextual Conflict Reasoning}, which performs visual-semantic conflict analysis across internal and external knowledge contexts; and (2) \emph{Correlation-Guided Encoding and Decoding}, featuring positional encoding compression for low-correlation statements and adaptive decoding using correlation-weighted conflict scoring.
Extensive evaluations on E-VQA, InfoSeek, and OK-VQA benchmarks demonstrate state-of-the-art performance, achieving absolute accuracy improvements of $3.3-6.4\%$ over competing methods.

\noindent \textbf{Limitations and Future Work.} 
One limitation of our method is its requirement for explicit externalization of model knowledge prior to performing visual-centric conflict reasoning. Ideally, after receiving external knowledge contexts, the model should implicitly identify and resolve conflicts between internal and external knowledge, necessitating robust reasoning capabilities. In future work, we will explore integrating multimodal reasoning into the KB-VQA task to further enhance response accuracy.

%% file: sec/X_suppl.tex
\clearpage
\setcounter{page}{1}
\renewcommand\thesection{\Alph{section}}
\setcounter{section}{0}

\noindent This appendix presents additional materials and results. First, we detail our prompts in experiments in Sec.~\ref{sec:Prompts Details} to enhance comprehension. Then, we present additional experiments in Sec.~\ref{Additional Experiments}. Finally, qualitative results are presented in Sec.~\ref{Qualitative Results}.
\section{Prompts Details in CC-VQA}
\label{sec:Prompts Details}
\textbf{Prompt for Parametric Context Generation:}
\begin{tcolorbox}
    Here is the question: \textbf{\{Question\}} Please describe the image about the question by your own Knowledge.
\end{tcolorbox}

\noindent \textbf{Prompt for Disambiguated Question:}
\begin{tcolorbox}
    Please refer to the given image and rewrite the question so that entities and attributes are explicit, pronouns are disambiguated, and the wording is more specific. Keep the original intent and scope unchanged; do not add new information. Only output the rewritten question wrapped in $<$question$>$$<$/question$>$. Original question: \textbf{\{Question\}}
\end{tcolorbox}
\noindent \textbf{Prompt for Visual Rationale Extraction:}
\begin{tcolorbox}
    Here is the question: \textbf{\{Question\}}, Here is the selected section:\textbf{\{section\}}, Give your answer and put the feature or reason for the asnwer related to the image in \textcolor{teal}{$<$reason$>$} \textcolor{teal}{$<$/reason$>$}.
\end{tcolorbox}

\noindent \textbf{Prompt for Visual-Centric Conflict Analysis:}
\begin{tcolorbox}
Here is the question: \textbf{\{Question\}}. Below are the reasons supporting the answer derived from the retrieved information: \textbf{\{Reasons text\}}. ldentify the key features that distinguish the aforementioned categories and extract the features that need attention in \textcolor{teal}{$<$reason$>$} \textcolor{teal}{$<$/reason$>$} without detailed illustration or answer.
\end{tcolorbox}

\noindent \textbf{Prompt for Correlation-Aware Position Encoding:}
\begin{tcolorbox}
Here is the question: \textbf{\{Question\}}. Here is the feature to focus on: \textcolor{teal}{$<$reason$>$}\{Features\} \textcolor{teal}{$<$/reason$>$}. Here is the retrieved information: \textbf{\{Retrieved Information\}}. Short Answer:
\end{tcolorbox}

\section{Additional Experiments}
\label{Additional Experiments}
\subsection{Ablation Study of Compression Ratio $\tau$}
In this section, we present an ablation study on the choice of $\tau$ in our Correlation-Aware Positional Encoding method. Specifically, $\tau$ denotes the percentage of sentences with the lowest correlation scores selected for compression. We conducted comparative experiments on a subset of 10K samples from InfoSeek.  
The results demonstrate that the accuracy progressively increases as more low-correlation sentences are compressed. Based on \textbf{Observation 2} in the paper, we set $\tau=75\%$, which compresses the bottom $75\%$ of sentences ranked by correlation score during position encoding. This indicates that compressing low-correlation sentences enables the model to better focus on and utilize high-correlation sentences that are most likely to contain answers.
\begin{table}[h]
\centering
\caption{\textbf{Ablation of Compression Ratio $\tau$.}}
\label{tab:RL}
\begin{tabular}{lc  cc cc}
\toprule
Method &\textbf{$\tau$} & Accuracy \\ \midrule
 & 25\% & 43.5 \\
Qwen2.5-VL-7B & 50\% & 44.7 \\
 & 75\% & 45.0 \\
\bottomrule
\end{tabular}
\end{table}
\vspace{-2pt}

\subsection{Generalization of CC-VQA}
We validated the effectiveness and generalizability of our method by testing it on large-scale KB with 100M entries using Qwen3-VL-8B. The results demonstrate that our method still achieves a significant improvement of 3.1\% (47.7\% $\to$ 50.8\%) on a stronger model. Additionally, we evaluated the impact of retrieval size on the results. Compared to the default top-3, the method can achieve an additional 1\% improvement when using top-5 retrieval.
\begin{table}[h]
\vspace{-0.3cm}
\centering
\caption{\textbf{Generalization across Models and Retrieval Size.}}
\footnotesize 
\setlength{\tabcolsep}{2.5pt} 
\renewcommand{\arraystretch}{0.8}

\begin{minipage}[b]{0.44\linewidth} 
    \centering
    \begin{tabular}{l|c|c}
    \toprule
    Method & Model & Acc. \\
    \midrule
    Vanilla RAG & Qwen3-VL-8B & 47.7 \\
    CC-VQA  & Qwen3-VL-8B & 50.8 \\
    \bottomrule
    \end{tabular}
\end{minipage}
\hfill
\begin{minipage}[b]{0.5\linewidth}
    \centering
    \begin{tabular}{l|c|c}
    \toprule
    Method & Retrieval Size & Acc. \\
    \midrule
    CC-VQA & top-3 & 50.8 \\
    CC-VQA & top-5 & 51.8 \\
    \bottomrule
    \end{tabular}
\end{minipage}
\vspace{-0.3cm}
\end{table}

\subsection{Comparison with Thinking Model}
Although CC-VQA involves multiple calls to the VLM model, each module uses the \textbf{same} model. To address concerns about test-time scaling, we compared it with different models. As shown in the table, our method achieved the highest accuracy (50.8\%). Compared to using a stronger model (Qwen3-VL-8B-Thinking), our approach consumes fewer output tokens (192 vs. 817) and has lower latency (8.94s vs. 11.79s), but achieves higher accuracy. This confirms that our performance improvement stems from the proposed conflict mitigation methods.
\noindent
\begin{minipage}{0.5\textwidth}
    \scriptsize
        \captionof{table}{\textbf{Comparison with Thinking Model.}} 
    \renewcommand{\arraystretch}{1}
    \begin{tabular}{c|c|c|c|c}
        \toprule
        Method & Model & Out Tok. & Latency (s) & Acc. (\%)  \\
        \midrule
        Vanilla RAG & Qwen3-VL-8B-Instruct & 75.74 &5.2& 47.7 \\
        Vanilla RAG & Qwen3-VL-8B-Thinking  &817.93&11.79& 48.8 \\
        CC-VQA & Qwen3-VL-8B-Instruct & 192.90 &8.94& 50.8 \\
        \bottomrule
    \end{tabular}
    \label{tab:frame_set} 
\end{minipage}

\subsection{Inference Time}
In this section, we evaluate the inference time of our method. Specifically, we select a subset of 10K samples from InfoSeek for this evaluation. As shown in Table~\ref{tab:Inference Time}, compared to the CoCoA method, our approach benefits from token compression, resulting in lower per-sample inference time. Here, ``s/k'' denotes the time in seconds required to process 1K samples.

\begin{table}[h]
\centering
\caption{\textbf{Inference Time.}}
\label{tab:Inference Time}
\resizebox{0.5\textwidth}{!}{
\begin{tabular}{lc  cc cc}
\toprule
Method &Model & Inference Time (s/k) \\ \midrule
CoCoA & Qwen2.5-VL-7B & 713.3  \\
CC-VQA \textbf{(Ours)} & Qwen2.5-VL-7B & 616.4 \\
\bottomrule
\end{tabular}}
\end{table}
Compared to Wiki-PRF, our method achieves comparable latency (8.94s vs. 8.77s) with only one additional forward pass (6 vs. 5), while remaining entirely training-free. With 76 GB of A800 GPU memory usage, our CC-VQA achieves an accuracy of 45.1\% on InfoSeek, demonstrating its powerful capabilities.

\begin{table}[htbp]
\centering
\caption{\textbf{Comparison with Wiki-PRF.}} 
\label{tab:efficiency}
\scriptsize
\setlength{\tabcolsep}{1pt}
\renewcommand{\arraystretch}{1}
\begin{tabular}{l|ccc|c|c}
\toprule
\textbf{Method} &\textbf{Total Time} & \textbf{Analysis} & \textbf{Generation}&\textbf{Gen. Finetuning} &\textbf{Acc.(\%)} \\
\midrule
Wiki-PRF (SOTA) &8.77s&3.30s&5.34s&\ding{52} &42.8\\
CC-VQA (Ours) &8.94s & 4.99s &3.95s&\ding{56} &45.1 \\
\bottomrule
\end{tabular}
\end{table}

\section{Qualitative Results}
\label{Qualitative Results}

\subsection{Illustration of VCCR}

The key idea of VCCR is to decompose the analysis of the model and retrieval contexts into multiple fine-grained subtasks (e.g., extraction and comparison), thereby constructing visual conflict reasoning to enhance problem-solving accuracy. To validate its effectiveness, we conducted both manual and MLLM-based verification. The result demonstrates that decomposed subtasks maintain high accuracy, while the VCCR module ultimately achieves overall accuracy of over 84\%.

\begin{table}[htbp]
\centering
\caption{\textbf{Analysis of VCCR.}} 
\label{tab:verifier}
\scriptsize
\setlength{\tabcolsep}{4pt}
\renewcommand{\arraystretch}{1}
\begin{tabular}{l|c|ccc|c}
\toprule
\textbf{Verifier} &\textbf{\#Samples} & \textbf{Generation} & \textbf{Extraction} & \textbf{Analysis} & \textbf{VCCR} \\
\midrule
Human  & 100 & 96\% & 77\% & 98\% & 91\% \\
MLLM-based & 10,000 & 93.87\% & 82.31\% & 95.55\% & 84.79\% \\
\bottomrule
\end{tabular}
\end{table}

\begin{figure}[h!]
    \vspace{-0.3cm}
    \centering
    \includegraphics[width=1\linewidth]{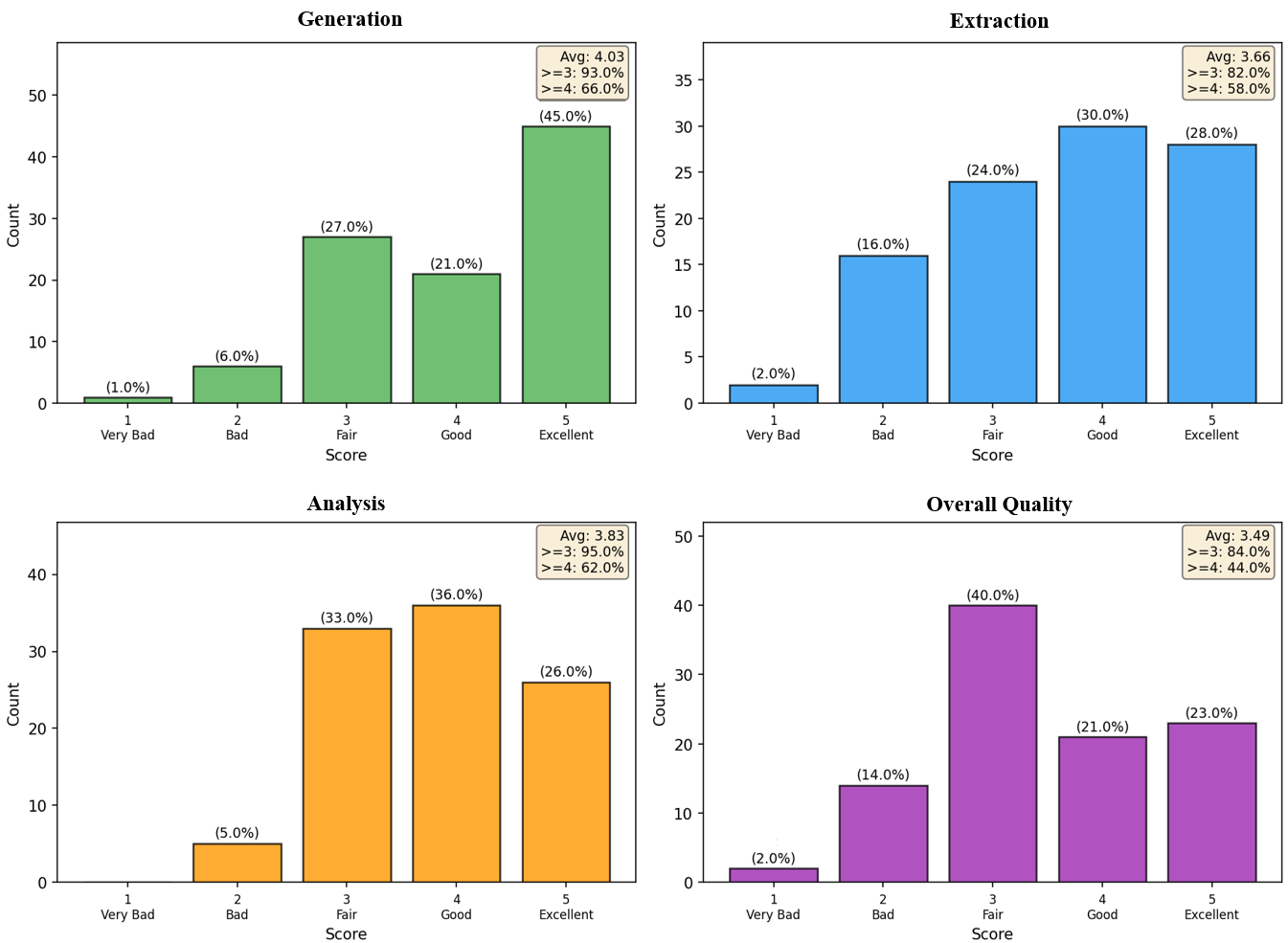}
    \caption{\textbf{Distribution of MLLM-based Verification for VCCR.}}
    \vspace{-0.5cm}
    \label{fig:distribution}
\end{figure}

Figure~\ref{fig:distribution} illustrates the score distribution across 10,000 evaluated samples. Our analysis indicates that the model's evaluation criteria are comparatively more stringent than those employed by human annotators. Consequently, scores of 3 or higher can be reasonably interpreted as indicative of accurate assessments. This empirical distribution further substantiates the validity and methodological soundness of the proposed VCCR framework.

\begin{figure*}[h!]
    \vspace{-0.3cm}
    \centering
    \includegraphics[width=1\linewidth]{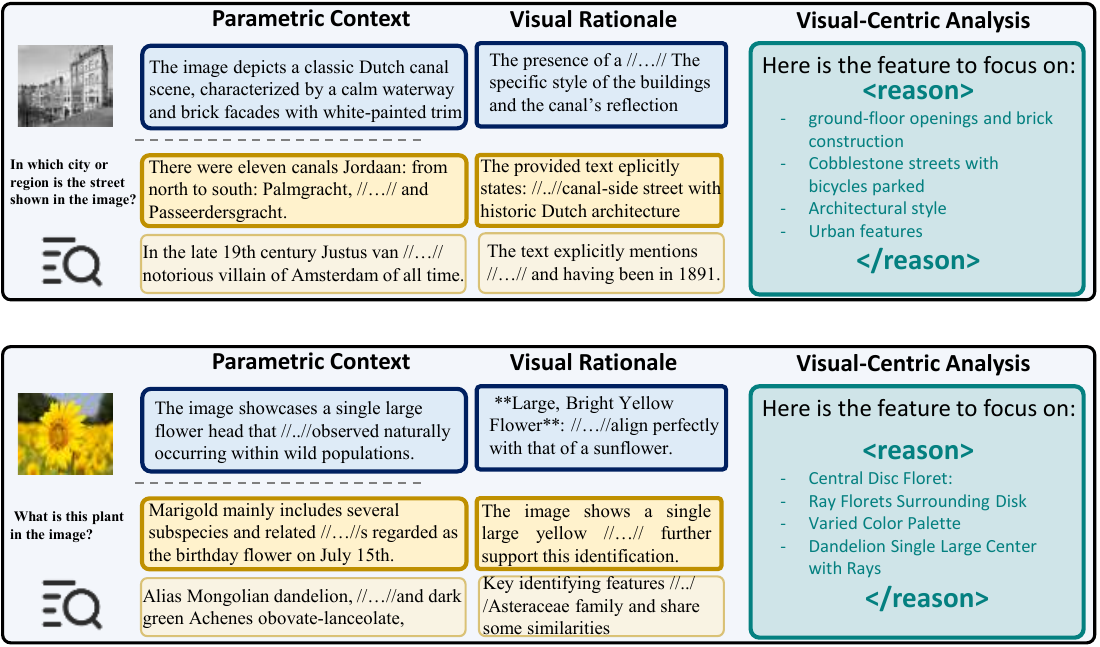}
    \caption{Detail Case of VCCR.}
    \vspace{-0.5cm}
    \label{fig:figure_task}
\end{figure*}

Two sample-level case study of VCCR is presented in Figure~\ref{fig:figure_task}. First, VCCR extracts parametric knowledge from the model based on visual inputs. It then executes reasoning on the retrieved data to isolate key features. In the final stage, all extracted features are consolidated to derive discriminative attributes, which serve as the basis for the ultimate reasoning process. The results demonstrate VCCR's ability to accurately capture fine-grained visual concepts and reasoning logic.

\subsection{More Comparison Cases}

\begin{figure*}[!t]
    \centering
    \includegraphics[width=1\linewidth]{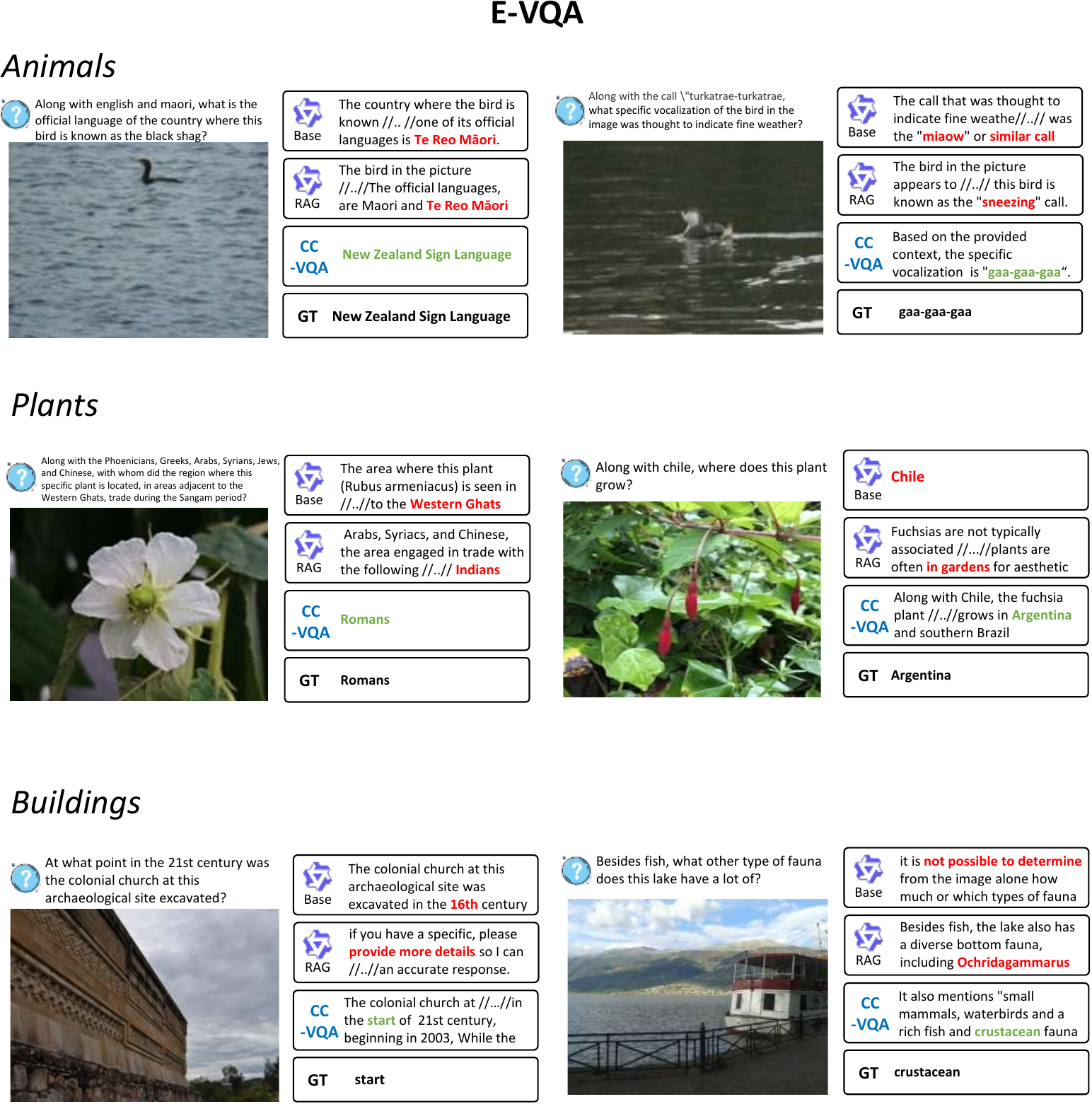}
    \caption{\textbf{More Cases of CC-VQA on E-VQA}. Qualitative results from Encyclopedic-VQA demonstrate that our method also enhances the model’s retrieval-augmented generation capabilities.}
    \label{fig: comparison cases}
\end{figure*}

In Figure~\ref{fig: comparison cases}, we present additional qualitative results of our method on Encyclopedic-VQA (E-VQA). Unlike the examples in the main text, these cases represent scenarios where both the base model and the standard RAG approach fail to produce correct answers. This further validates the effectiveness of sentence-level similarity for filtering relevant and accurate information. We specifically select three categories: animals, plants, and buildings. For each category, we show two types of questions: one indirectly related and one directly related to the image content.

\begin{figure*}[!t]
    \centering
    \includegraphics[width=1\linewidth]{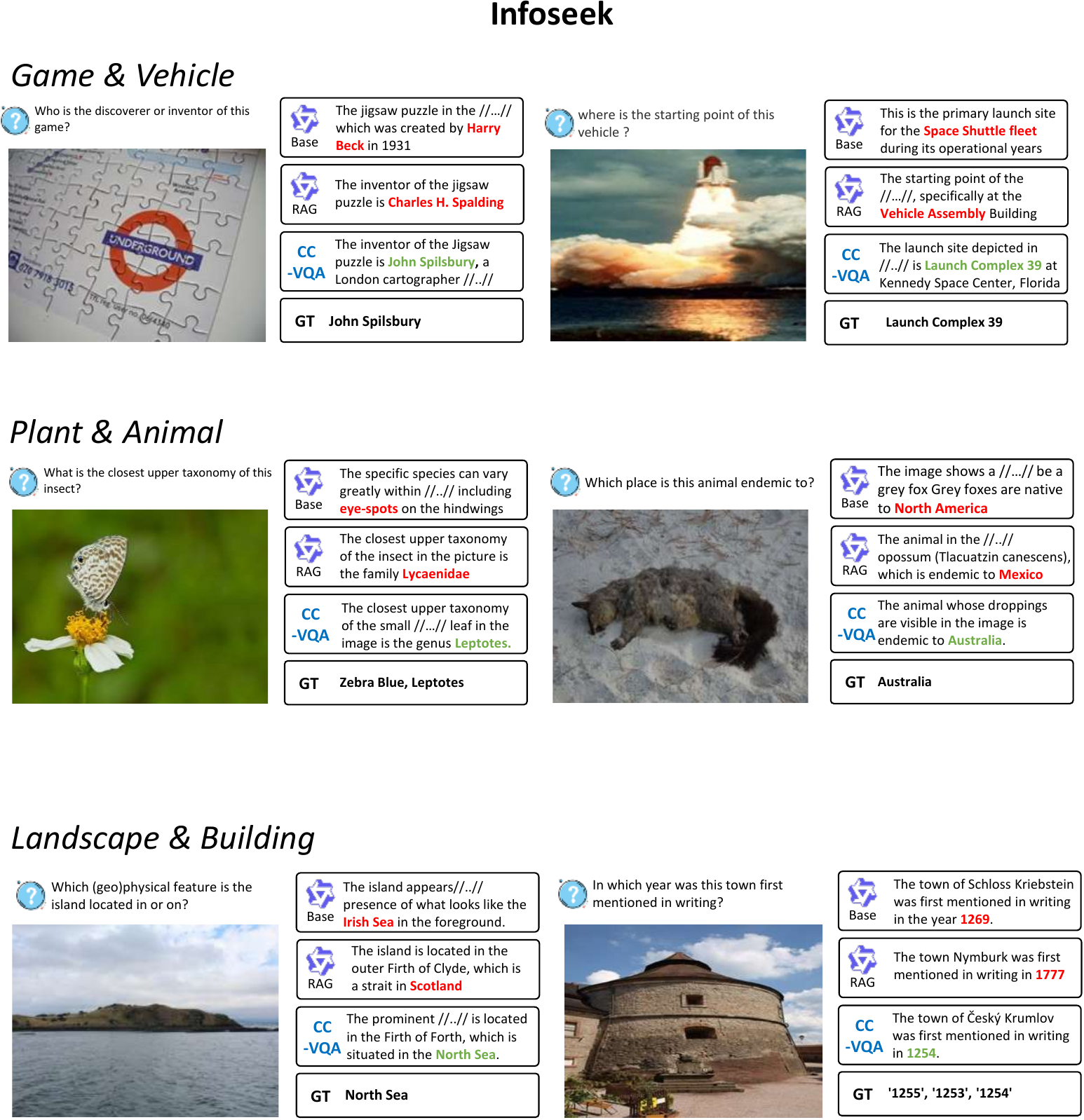}
    \caption{\textbf{More Cases of CC-VQA on Infoseek}. Qualitative results from Infoseek demonstrate that our method also enhances the model’s retrieval-augmented generation capabilities.}
    \label{fig: comparison cases1}
\end{figure*}

In Figure~\ref{fig: comparison cases1}, we present additional qualitative results of our method on Infoseek. These cases further enrich the scope of our evaluation by including examples such as the author of a jigsaw puzzle, the launch site of a rocket, and the time zone of a small town. The results demonstrate that our correlation computation effectively enhances the model’s ability to extract correct answers from retrieved information.

\subsection{Illustration of CC-VQA}
\begin{figure*}[!t]
    \centering
    \includegraphics[width=0.9\linewidth]{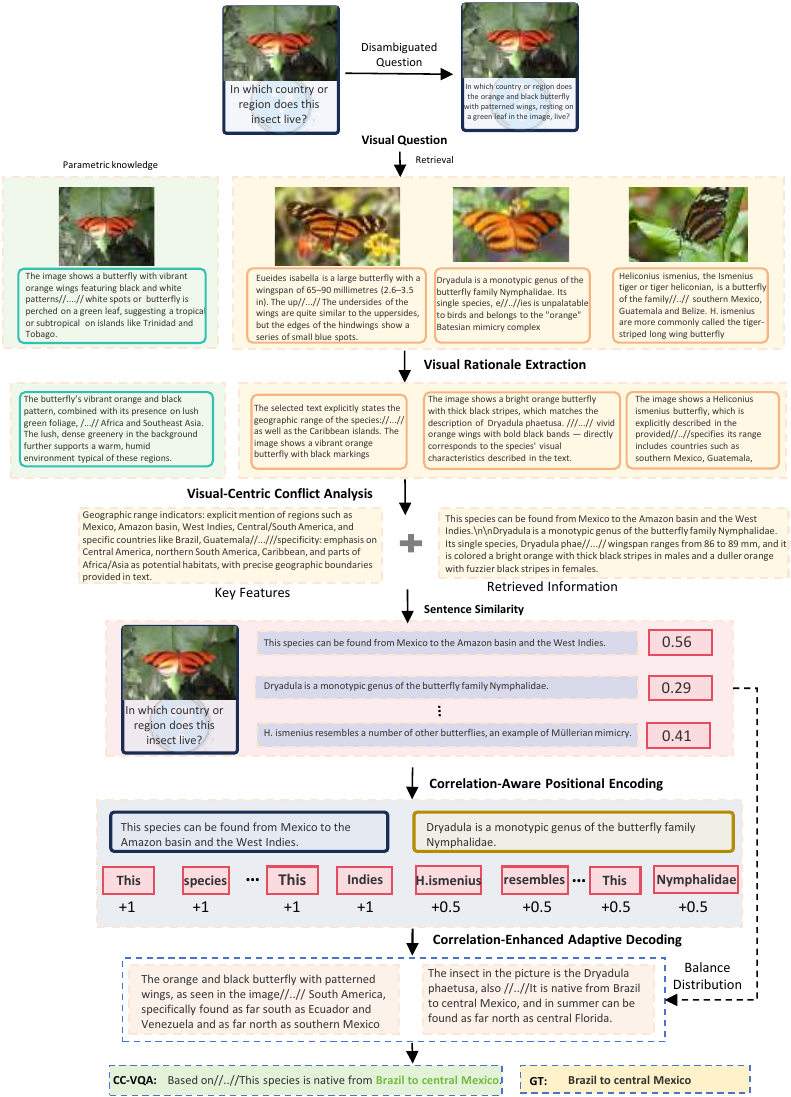}
    \caption{\textbf{Illustration of CC-VQA on E-VQA.}}
    \label{fig: full cases2}
\end{figure*}

\begin{figure*}[!t]
    \centering
    \includegraphics[width=0.9\linewidth]{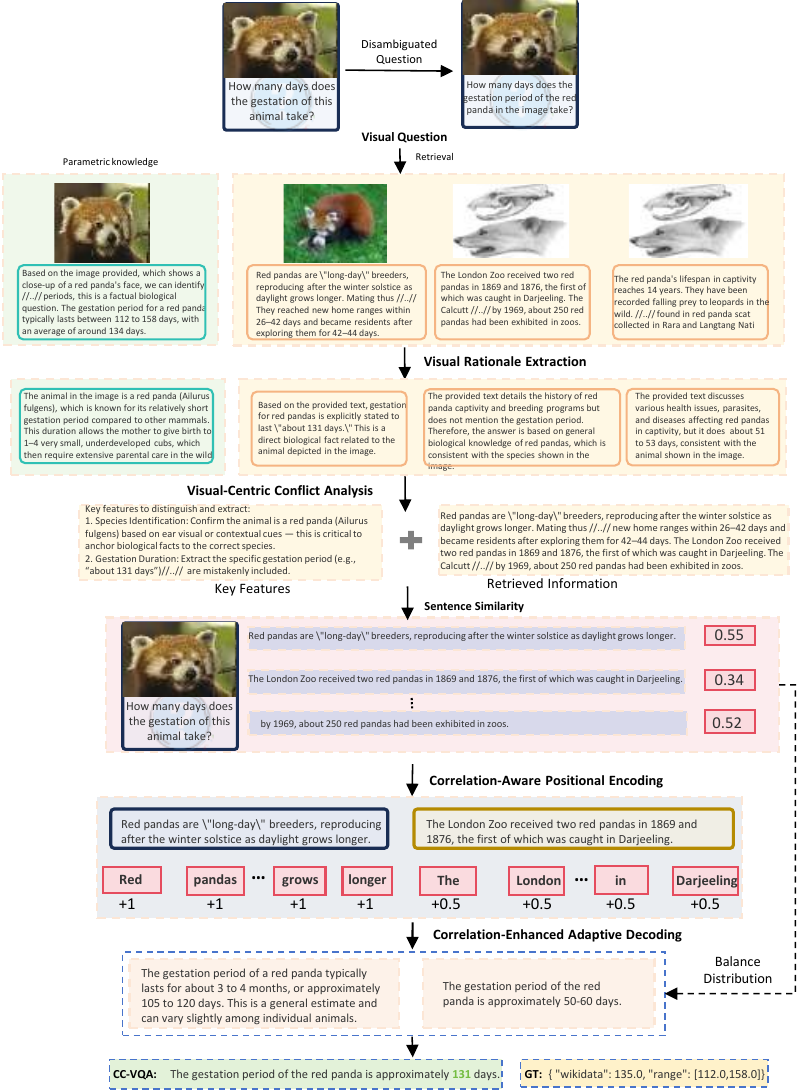}
    \caption{\textbf{Illustration of CC-VQA on Infoseek.}}
    \label{fig: full cases1}
\end{figure*}

In this section, we mainly show the case examples of CC-VQA. Figure~\ref{fig: full cases2} and Figure~\ref{fig: full cases1} illustrate complete inference pipelines on CC-VQA. Specifically, Figure~\ref{fig: full cases2} presents a location-related case from E-VQA, while Figure~\ref{fig: full cases1} shows a time-related case from InfoSeek.

